\documentclass{article}
\usepackage{iclr2027_conference,times}

\usepackage{amsmath,amsfonts,bm}

\def\eqref#1{equation~\ref{#1}}

\def\1{\bm{1}}

\DeclareMathAlphabet{\mathsfit}{\encodingdefault}{\sfdefault}{m}{sl}
\SetMathAlphabet{\mathsfit}{bold}{\encodingdefault}{\sfdefault}{bx}{n}

\iclrfinalcopy
\usepackage{hyperref,url,graphicx,booktabs,amsmath,amssymb,subcaption,caption,placeins,wrapfig,afterpage,multirow}
\usepackage{array}


\newcommand{\vina}[2]{#1\ {\tiny ($\pm$ #2)}}
\newcommand{\vinab}[2]{\textbf{#1}\ {\tiny ($\pm$ #2)}}
\newcolumntype{C}[1]{>{\centering\arraybackslash}m{#1}}

\title{Transferring Peptide–Target Interaction Patterns to Small Molecules by Joint Guidance for Diffusion and Flow Matching}
\author{
\href{https://openreview.net/profile?id=~Jian_Lee2}{\textit{Jian Lee}}$^{1}$ \quad
\href{https://openreview.net/profile?id=~Changyun_Cho1}{\textit{Changyun Cho}}$^{1}$ \quad
\href{https://openreview.net/profile?id=~Sun_Kim2}{\textit{Sun Kim}}$^{1,2,3,4}$\\[0.4em]
$^{1}$AIGENDRUG Co., Ltd., Seoul\\
$^{2}$Interdisciplinary Program in Artificial Intelligence, Seoul National University\\
$^{3}$Department of Computer Science and Engineering, Seoul National University\\
$^{4}$Interdisciplinary Program in Bioinformatics, Seoul National University
}

\begin{document}
\maketitle
\lhead{Preprint}

\begin{abstract}
Generating small molecules to target protein-protein interactions (PPIs) remains one of the most challenging yet high-reward frontiers in drug discovery. Unlike traditional small-molecule drugs that bind to deep, well-defined pockets, PPI interfaces often comprise broad, shallow binding surfaces. A natural strategy is to leverage known peptide–target interactions, as peptide ligands provide rich interface-level constraints. However, a major algorithmic challenge is that rigid atom-level conditioning on peptide structures can over-constrain the generative space and exclude valid small-molecule solutions. 
To address this challenge, we present PIFT (Peptide InterFace Transfer), a novel framework that transfers peptide–target interaction patterns to small molecules through joint sampling-time guidance of masked discrete diffusion and continuous flow matching. 
PIFT first maps peptide–target and small-molecule–target interfaces into a shared, target-conditioned interaction space, capturing transferable interface information across molecular modalities. The resulting peptide-derived reference then jointly steers building-block logits and 3D coordinates toward peptide-compatible molecules while preserving a strong adapted molecular prior.
Extensive experiments across diverse PPI targets show that the learned representation captures cross-modal interaction compatibility, while peptide guidance improves interface recovery and achieves strong docking performance and favorable molecular properties.
\end{abstract}

\section{Introduction}

Protein--protein interactions (PPIs) regulate diverse cellular processes and are broadly implicated in human disease, making them an important class of therapeutic targets \citep{greenblatt2024,nada2024}. Their interfaces, however, are typically broad, shallow, and weakly pocketed, with extended and heterogeneous surface geometries that make effective small-molecule engagement particularly challenging \citep{arkin2004,jin2014}. Recent structure-based generative models have substantially expanded the design space by conditioning molecular identity and three-dimensional geometry on protein structural context \citep{peng2022pocket2mol,guan2023targetdiff}, while newer synthesis-aware formulations additionally couple molecular generation to chemically feasible construction pathways \citep{cretu2025synflownet,seo2025rxnflow,shen2025cgflow}. These models can further incorporate reference-ligand information \citep{huang2024irdiff} or explicit protein--ligand interaction constraints \citep{zhung2024deepicl}, providing increasingly fine-grained control over generation. For PPIs, however, specifying an appropriate binding condition remains difficult: the relevant interface may extend beyond a compact ligand-binding site, leaving the desired interface constraint under-specified. This motivates PPI-specific methods that derive design cues directly from the target interaction interface.

A prominent class of PPI-specific methods derives molecular design constraints from interface hotspots. In particular, hotspot-derived pharmacophore conditioning represents a broad PPI interface through a compact spatial arrangement of selected energetically or functionally important features \citep{wells2007,sun2026hot2mol}. The effectiveness of such methods establishes the utility of interface-derived information for small-molecule design. Nevertheless, hotspot conditioning intentionally compresses the interface into a selected subset of spatial and chemical cues, limiting the extent to which broader multi-contact organization is retained.

An alternative is to draw design information from the interaction partner itself. Many PPIs are mediated by short peptide motifs, and bound peptides therefore provide structurally realized examples of how multiple contacts are organized across an extended target surface \citep{corbiverge2017,pelaygimeno2015,vanwier2025}. Recent peptide-conditioned generative approaches encode peptide and protein structures at the atomic level, often augmented with residue-level context, and inject these representations into molecular denoisers through equivariant or attention-based conditioning mechanisms \citep{he2025peptide2mol,yue2026pep2mol}.

However, transferring peptide-defined interfaces into small-molecule generation still poses two challenges.

\begin{itemize}
    \item \textbf{Structural mismatch across molecular modalities.}
    Peptides and small molecules realize binding through fundamentally different molecular organizations: peptides use residue-indexed backbone and side-chain architectures, whereas small molecules arrange heterogeneous functional groups on compact scaffolds. Relying on fine-grained atomic representations of peptide structure can over-constrain the space of valid small-molecule realizations. This structural mismatch poses a fundamental challenge for peptide-to-small-molecule translation.    
    
    \item \textbf{Guidance across discrete chemistry and 3D geometry.}
    Once the peptide interface is transferred into a shared representation, using it to control generation remains nontrivial. Synthesis-aware molecular generation involves both discrete building-block assembly and continuous three-dimensional geometry. Because these two components respond to conditioning in fundamentally different ways, incorporating peptide-derived constraints without distorting the pretrained molecular distribution remains challenging.
\end{itemize}

\begin{figure*}[t]
\centering
\includegraphics[width=\textwidth]{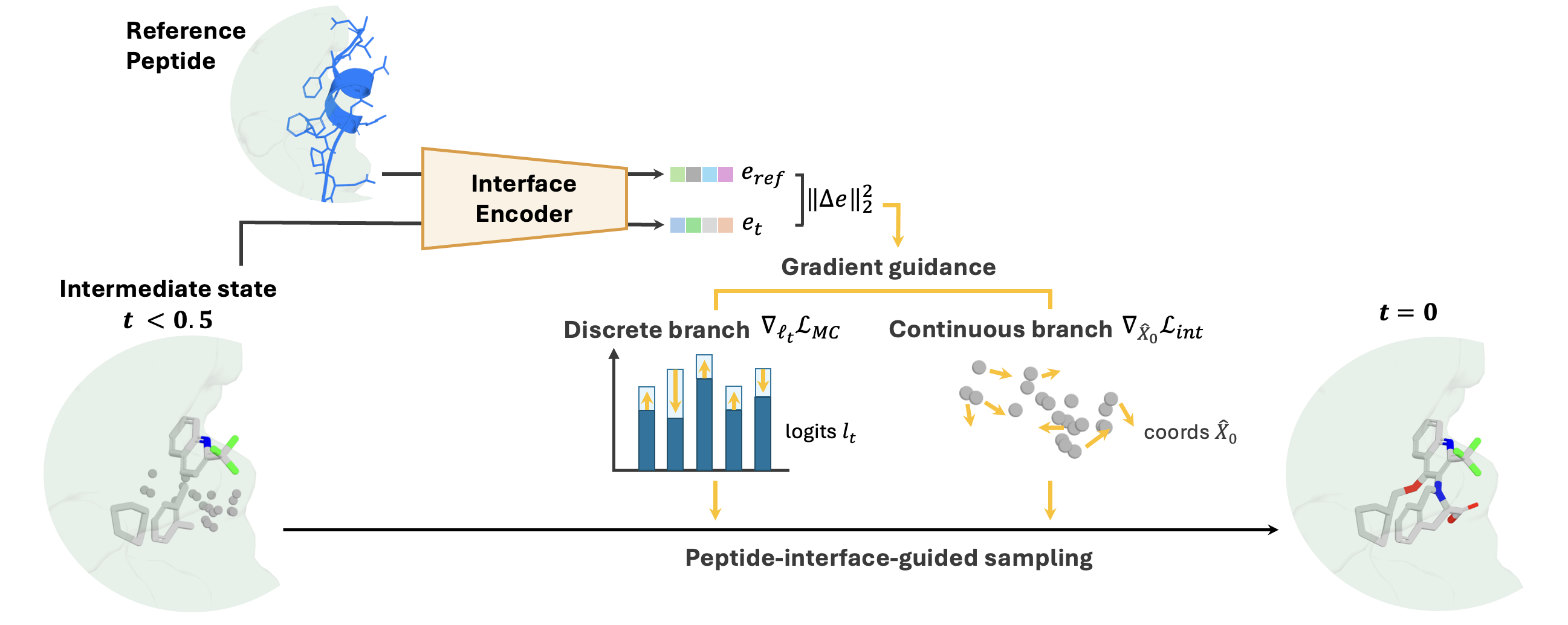}
\captionsetup{skip=10pt}
\caption{\textbf{Overview of PIFT molecular generation framework.} The interface encoder extracts representations from a reference peptide–target interface and an intermediate molecule–target state. The guidance objective derived from their embeddings jointly steers discrete building-block logits and continuous 3D coordinates during sampling.}
\label{fig:overview}
\vspace{-0.5em}
\end{figure*}

To address these challenges, we introduce PIFT, formulating peptide-to-small-molecule generation as transferring peptide--target interaction patterns through joint control of discrete molecular identity and continuous 3D geometry. PIFT learns a shared target-conditioned interaction space, adapts a pretrained synthesis-aware generator toward a target-specific PPI-oriented prior, and uses the peptide interface as a sampling reference (Figure~\ref{fig:overview}).

Our contributions are:
\begin{itemize}
    \item We introduce a novel target-conditioned cross-modal representation that transfers peptide–target interaction patterns to small molecules through shared interaction types and three-dimensional geometry.
    \item We develop a joint peptide-reference guidance mechanism that applies a single interface objective to both masked discrete diffusion over building blocks and continuous flow matching over 3D geometry, enabling coordinated control of chemistry and pose.
    \item Across diverse PPI targets, we show that the learned representation captures cross-modal interaction compatibility and that PIFT effectively recovers the peptide interface while achieving strong docking performance and molecular properties.
\end{itemize}

\section{Background and Related Work}

\subsection{Pharmacophore and Peptidomimetic Design for PPI Inhibitors}

Peptidomimetic design replaces a peptide or protein interaction motif with a chemically distinct scaffold while preserving the key determinants required for binding. Pharmacophore-based approaches provide a natural abstraction for this transfer by representing interaction-relevant functional features and their spatial organization independently of the underlying molecular topology \citep{wells2007,voets2012,pelaygimeno2015,qvit2017}. This abstraction motivates our use of a shared pharmacophore representation to compare peptide and small-molecule interfaces under the same target context.

\subsection{Discrete Synthesis-Aware Molecular Generation}

\paragraph{Masked discrete diffusion.}
Let $x_0$ denote a categorical object represented over a vocabulary that includes an absorbing mask token. Discrete denoising diffusion progressively corrupts $x_0$ according to a cumulative transition matrix $\bar Q_t$ and trains a reverse model to predict the clean categorical state \citep{austin2021d3pm}:
\begin{equation}
q(x_t\mid x_0)=\operatorname{Cat}(x_t;x_0\bar Q_t),
\qquad
p_\theta(x_0\mid x_t,t).
\label{eq:masked-diffusion}
\end{equation}
With absorbing-mask corruption, generation proceeds by iteratively resolving masked variables.
The reverse transition can be written in terms of the predicted clean-state posterior as
\begin{equation}
p_\theta(x_s\mid x_t)=\sum_{x_0}q(x_s\mid x_t,x_0)p_\theta(x_0\mid x_t),\qquad s<t,
\end{equation}
which exposes the clean-state prediction as a natural point for sampling-time categorical guidance. 
In synthesis-aware molecular generators, these categorical variables can represent building blocks and reaction choices, grounding generation in synthesizable chemical space.

\subsection{Sampling-Time Guidance in Continuous and Discrete Spaces}

\paragraph{Continuous gradient guidance.}
Sampling-time guidance controls a pretrained generative trajectory through an external objective without retraining the full model \citep{dhariwal2021,bansal2024universal}. If a differentiable loss $\mathcal L(\hat x_0,c)$ is evaluated on the predicted clean state $\hat x_0=\hat x_{0,\theta}(x_t)$, the reverse score can be modified as
\begin{equation}
\widetilde s_\theta(x_t,t;c)
=s_\theta(x_t,t)-\lambda_t\nabla_{x_t}
\mathcal L(\hat x_{0,\theta}(x_t),c).
\label{eq:loss-guidance}
\end{equation}
This illustrates the general principle of differentiating a downstream objective through a denoised prediction; our geometric guidance applies the resulting correction directly to the predicted clean coordinates.

\paragraph{Guidance for categorical states.}
Discrete diffusion poses a distinct challenge for sampling-time guidance because its categorical states are not directly differentiable. GILC addresses this difficulty by building on straight-through Gumbel--Softmax \citep{jang2017} to obtain gradients through discrete samples and applying them as Jacobian-free corrections to clean-prediction logits \citep{dou2026gilc}. This provides the basis for our discrete guidance branch, which applies peptide-reference gradients to building-block logits during sampling.

\subsection{Reward-Based Adaptation of Generative Priors}
Reward tilting adapts a pretrained generative distribution toward high-reward regions while retaining support from the original prior. For masked discrete diffusion, Discrete Tilt Matching (DTM) realizes this principle by matching reward-tilted local unmasking posteriors at intermediate states, avoiding direct sequence-level likelihood optimization \citep{chen2026dtm, wang2025drakes}. We use DTM to construct the PPI-oriented molecular prior described in Section~\ref{sec:prior_adaptation}.

\section{Method}

\subsection{Problem Formulation}

For a target protein with a bound reference peptide, let $C_{\mathrm{pep}}$ and $T$ the denote the target-facing peptide and receptor interface representations, respectively. We generate a synthesizable molecule and pose $x=(G,\mathbf X)$, where $G$ denotes the reaction graph, including building-block identities and coupling reaction (and centers), and and $\mathbf{X} \in \mathbb{R}^{N \times 3}$ contains the corresponding 3D coordinates.
The desired output should recover a target-facing interaction pattern compatible with that of the reference peptide.

\subsection{Transferring Peptide--Target Interaction Patterns to Small Molecules}
\label{sec:interaction_transfer}

\subsubsection{Shared Interface Representation}

\paragraph{Dataset construction.}
\label{sec:dataset}
We construct the cross-modal interface corpus from peptide--protein and small-molecule--protein complexes in \textbf{PepBDB}, \textbf{CrossDocked2020}, \textbf{2P2Idb v2}, and \textbf{DLiP-PPI} \citep{wen2019pepbdb,francoeur2020crossdocked,basse2016twop2i,ikeda2023dlip}. Complexes are aligned within a common target coordinate frame, and splits are defined by UniProt accession to prevent target leakage. Detailed processing and corpus statistics are provided in Appendix~\ref{app:dataset}.

\paragraph{Shared Representation.}
We represent peptide and small-molecule interfaces through target-facing interaction features extracted with ProLIF \citep{bouysset2021prolif}. Peptide and small-molecule interfaces share the same eight-type vocabulary: hydrogen-bond donor and acceptor, aromatic, hydrophobic, cationic, anionic, halogen-bond, and metal-binding features. Each binder-side interface is represented as a typed three-dimensional point cloud
$C=\{(\mathbf{x}_i,c_i,\mathbf{d}_i,v_i)\}_{i=1}^{N}$,
where $\mathbf{x}_i\in\mathbb{R}^3$ denotes the feature coordinate, $c_i$ the interaction type, $\mathbf{d}_i$ an optional interaction direction, and $v_i\in\{0,1\}$ a validity mask. The corresponding target cloud $T$ uses complementary receptor-side interaction types associated with observed contacts, with each point placed at the closest participating target atom for its residue-pair/type combination.

\subsubsection{Target-Conditioned Cross-Modal Interface Encoder.}
\label{sec:pharmacophore}

\paragraph{Encoder architecture.}
Peptide and small-molecule interfaces are processed by a shared encoder. Each interface point is initialized from its interaction type and SE(3)-invariant geometric features, followed by pairwise message passing and masked pooling:

\vspace{-0.9em}
\begin{align}
\mathbf h_i^{(0)}
&=
f_{\mathrm{point}}
\!\left(
[\mathrm{Emb}(c_i),\boldsymbol\rho_i]
\right)v_i,
\\
\mathbf h_i^{(1)}
&=
\mathrm{LN}\!\left(
\mathbf h_i^{(0)}
+
\frac{1}{|\mathcal N(i)|}
\sum_{j\in\mathcal N(i)}
f_{\mathrm{msg}}
\!\left(
\mathbf h_i^{(0)},
\mathbf h_j^{(0)},
\boldsymbol\rho_{ij}
\right)
\right)v_i,
\\
\mathbf p
&=
\frac{\sum_i v_i\mathbf h_i^{(1)}}
{\sum_i v_i}.
\end{align}
where $\mathcal N(i)=\{j:v_j=1,\ \|\mathbf x_i-\mathbf x_j\|_2<r_c\}$, and $\boldsymbol\rho_i$ and $\boldsymbol\rho_{ij}$ denote translation- and rotation-invariant geometric scalars.
A separate target encoder produces $\mathbf t=E_{\mathrm{target}}(T)$, which is mapped to FiLM \citep{perez2018film} parameters and applied to the pooled representation before final projection:

\vspace{-0.9em}
\begin{equation}
\begin{split}
(\boldsymbol\gamma,\boldsymbol\beta)
&= f_{\mathrm{FiLM}}(\mathbf t),\\
E_\phi(C,T)
&= f_{\mathrm{out}}
\left(
\boldsymbol\gamma \odot \mathbf p + \boldsymbol\beta
\right)
\in \mathbb R^{64}.
\end{split}
\label{eq:encoder}
\end{equation}

The FiLM modulation is initialized at $\boldsymbol\gamma=\mathbf{1}$ and $\boldsymbol\beta=\mathbf{0}$, so that target conditioning initially acts as an identity transformation.

\paragraph{Phase I: modality-shared geometric pretraining.}
We first pretrain target-agnostic $E_\phi(C)$ on balanced peptide and small-molecule batches. Following the training scheme of PepMimic \citep{kong2026pepmimic}, each cloud yields two overlapping views $C_a,C_b$ and a spatially disjoint view $C_n$:
\begin{equation}
\mathcal L_{\mathrm{partial}}
=\|E_\phi(C_a)-E_\phi(C_b)\|_2
+\mathrm{ReLU}\!\left(m_{\mathrm{p}}-\|E_\phi(C_a)-E_\phi(C_n)\|_2\right).
\label{eq:partial}
\end{equation}
The objective brings overlapping views together while separating spatially disjoint views, providing a target-agnostic initialization for Phase II.

\paragraph{Phase II: target-conditioned cross-modal alignment.}
Starting from the Phase-I encoder, we introduce the target encoder and FiLM conditioning and train peptide queries with same-target molecular examples. All examples in a query group share target representation $\mathbf t_q$. For query $q$, positive $p$, and negative $n$,
\begin{equation}
\mathcal L_{\mathrm{rank}}=\mathbb E_{q,p,n}\!\left[\operatorname{softplus}\!\left(\frac{\|\mathbf e_q-\mathbf e_p\|_2-\|\mathbf e_q-\mathbf e_n\|_2+m_{\mathrm r}}{\tau_{\mathrm r}}\right)\right],
\label{eq:rank}
\end{equation}
where $\mathbf e_u=E_\phi(C_u,T_q)$. We regularize the FiLM scale and discourage collapse of target representations:
\begin{equation}
\begin{gathered}
\mathcal L_\gamma
= \mathbb E\|\boldsymbol\gamma-\mathbf 1\|_2^2,
\qquad
\mathcal L_{\mathrm{div}}
= \mathbb E_{k\neq l}
\left[
\mathrm{ReLU}\!\left(
\cos(\mathbf t_k,\mathbf t_l)-\tau_{\mathrm{div}}
\right)^2
\right],
\\[2pt]
\mathcal L_{\mathrm{enc}}
=
\mathcal L_{\mathrm{rank}}
+\lambda_\gamma\mathcal L_\gamma
+\lambda_{\mathrm{div}}\mathcal L_{\mathrm{div}}.
\end{gathered}
\label{eq:encoder_objective}
\end{equation}
The resulting space places peptide and small-molecule interfaces with compatible target-facing interactions closer together under the same target context.

\subsection{Guiding Molecular Generation toward Peptide–Target Interfaces}

\begin{figure}[!t]
    \centering
    \includegraphics[
        width=\linewidth
    ]{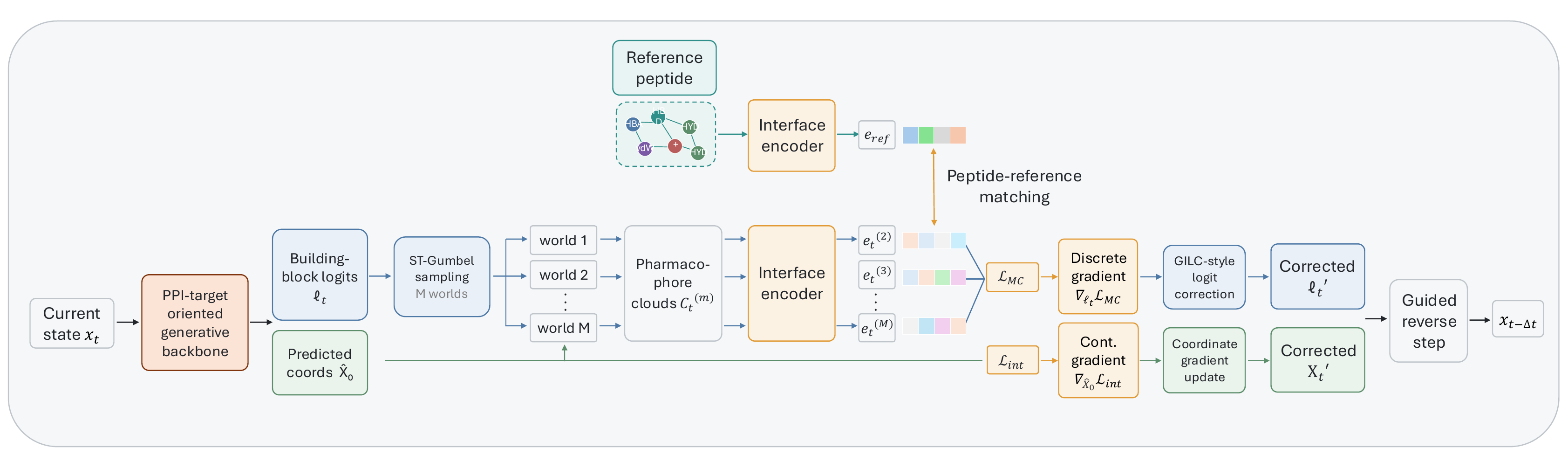}
    \caption{
        \textbf{Branch-specific mechanism of peptide-reference guidance.}
    }
    \label{fig:main_guidance}
    \vspace{-0.5em}
\end{figure}

\subsubsection{PPI-Oriented Molecular Prior Adaptation}
\label{sec:prior_adaptation}
We use SynCoGen \citep{rekesh2026syncogen} as the synthesis-aware generative backbone and adapt its pretrained molecular prior toward a target-specific PPI-oriented distribution using Discrete Tilt Matching (DTM).
Given the pretrained prior $p_0(x)$, DTM defines the reward-tilted distribution
\begin{equation}
p_R(x \mid T)
=
\frac{1}{Z_R(T)}
p_0(x)\exp\!\left(\beta R_{\mathrm{PPI}}(x,T)\right),
\qquad
Z_R(T)
=
\mathbb{E}_{p_0}
\left[
\exp\!\left(\beta R_{\mathrm{PPI}}(x,T)\right)
\right].
\label{eq:dtm-tilt}
\end{equation}

DTM then adapts the discrete denoiser toward this reward-tilted distribution through local unmasking-posterior matching. 
Details of the reward and optimization are provided in Appendix~\ref{app:dtm}.

\subsubsection{Joint Guidance of Discrete Diffusion and Continuous Flow Matching}
\label{sec:guidance}

\paragraph{Shared interface objective.}
After target adaptation, the generator and interface encoder are frozen, and reference embedding $\mathbf e_{\mathrm{ref}}=E_\phi(C_{\mathrm{pep}},T)$ is computed once.
At reverse step $t$, the backbone predicts clean-state building-block logits $\boldsymbol{\ell}_t$ and a denoised coordinate estimate $\hat{\mathbf X}_0$.
Let $\bar E_t$ denote the committed graph connectivity, which is held fixed during guidance.
For an interface realization $C$, we define the shared peptide-reference objective
$\mathcal L_{\mathrm{int}}(C)
=
\left\|
E_\phi(C,T)-\mathbf e_{\mathrm{ref}}
\right\|_2^2$.
The discrete and continuous branches optimize this same objective through different differentiable realizations of molecular identity and geometry (Figure~\ref{fig:main_guidance}).

\paragraph{Discrete building-block guidance.}
For each unresolved building-block node, we construct $M$ differentiable categorical
realizations from the clean-state logits $\ell_t$ using straight-through
Gumbel--Softmax \citep{jang2017}:
\begin{equation}
\mathbf y_n^{(m),\mathrm{soft}} = \operatorname{softmax} \left( \frac{\ell_{t,n}+\boldsymbol{\eta}_n^{(m)}}{\tau_G} \right), \qquad \mathbf y_n^{(m),\mathrm{ST}} = \operatorname{sg} \left( \mathbf y_n^{(m),\mathrm{hard}} - \mathbf y_n^{(m),\mathrm{soft}} \right) + \mathbf y_n^{(m),\mathrm{soft}},
\label{eq:st-gumbel}
\end{equation}
where $\boldsymbol{\eta}_n^{(m)}$ is i.i.d. Gumbel noise and
$\operatorname{sg}(\cdot)$ denotes stop-gradient. Committed building blocks
are shared across worlds, while unresolved nodes may realize different
candidate identities.

Each world induces the pharmacophore realization
$C_t^{(m)}=\Phi\!\left(\mathbf y^{(m),\mathrm{ST}},\bar E_t,\hat{\mathbf X}_0\right)$,
yielding the Monte Carlo interface objective 
\vspace{-0.5em}
\begin{equation}
\mathcal L_{\mathrm{MC}} = \frac{1}{M} \sum_{m=1}^{M} \left\| E_\phi(C_t^{(m)},T) - \mathbf e_{\mathrm{ref}} \right\|_2^2.
\label{eq:mc-interface}
\end{equation}
Its gradient $\mathbf g_\ell=\nabla_{\ell_t}\mathcal L_{\mathrm{MC}}$ is
converted into a GILC-style clean-logit correction \citep{dou2026gilc}. The
gradient is restricted to unresolved nodes with valid reaction-supported
realizations, then normalized, clipped, and smoothed.

We apply the guidance only during the later reverse steps using the schedule
\begin{equation}
r_{\mathrm{disc}}(t) = \left[ \max \left( 0, \frac{t_{\mathrm{start}}-t} {t_{\mathrm{start}}} \right) \right]^{p_{\mathrm{disc}}}.
\label{eq:discrete-schedule}
\end{equation}
With a fixed maximum discrete guidance scale
$\gamma_{\max}$, the corrected clean logits are
\begin{equation}
\boldsymbol{\ell}_{t,n}
\leftarrow
\boldsymbol{\ell}_{t,n}
-
\gamma_{\max}
r_{\mathrm{disc}}(t)
\hat{\mathbf g}_{\ell,n}.
\end{equation}

\paragraph{Continuous geometric guidance.}
Because the interface cloud is differentiable with respect to
$\hat{\mathbf X}_0$, we directly compute
$\mathbf g_X=\nabla_{\hat{\mathbf X}_0}\mathcal L_{\mathrm{int}}(C_t)$.
After removing the global translation component and applying normalization
and clipping, we update
\vspace{-0.4em}
\begin{equation}
\hat{\mathbf X}_0\leftarrow\hat{\mathbf X}_0
-t\,\gamma_C(t)\hat{\mathbf g}_{X},\qquad
\gamma_C(t)=\gamma_{\max}\left[\max\!\left(0,\frac{t_{\mathrm{start}}-t}{t_{\mathrm{start}}}\right)\right]^{p_C}.
\label{eq:coordinate-guidance}
\end{equation}
This applies the same peptide-reference objective directly to 3D geometry.

We apply RMS normalization, clipping, and adaptive moment smoothing to the
guidance gradients; implementation details are provided in
Appendix~\ref{app:gradient-stabilization}.

\section{Experiments}

\subsection{Experimental Setup}

\paragraph{Targets and sampling protocol.}
We evaluate our method and external baselines on 10 predefined PPI targets spanning distinct interface geometries and target families. For the guidance ablation, we compare PIFT with the same target-adapted molecular prior without peptide-reference guidance (Off), holding target, prior checkpoint, sampling budget, and random seeds fixed whenever supported.

\paragraph{Baselines.}
We compare against representative alternatives covering structure-based 3D generation, synthesis-aware molecular generation, and peptide/PPI-guided design. \textbf{TargetDiff}, \textbf{MolCRAFT}, and \textbf{FLOWR} provide strong structure-conditioned diffusion or flow-based baselines \citep{guan2023targetdiff,qu2024molcraft,cremer2026flowr}; \textbf{CGFlow}, \textbf{SynCoGen}, and \textbf{SynGA} represent synthesis-aware or search-based generation \citep{shen2025cgflow,rekesh2026syncogen,lo2026synga}. CGFlow is evaluated in a target-specific setting, with a separate model trained for each target. We further include \textbf{Hot2Mol} as a hotspot-guided PPI inhibitor generator and \textbf{Peptide2Mol} as a peptide-conditioned peptidomimetic diffusion model \citep{sun2026hot2mol,he2025peptide2mol}. Architectural details are provided in Appendix~\ref{sec:baseline_details}.

\paragraph{Metrics.}
While peptidomimetic design emphasizes preservation of key binding determinants across distinct scaffolds \citep{pelaygimeno2015}, standard docking and molecular-quality metrics do not assess recovery of the reference peptide’s target-facing interaction pattern. 
We therefore introduce \textbf{Interface Agreement (IFA)}, which compares a
reference peptide $P$ and molecule $L$ through residue-wise target interaction
profiles. For binder $B\in\{P,L\}$, target residue $r$, and ProLIF interaction
type $t$, we define
\[
\mathbf u_B(r,t)=
\begin{cases}
\exp\!\left[-d_{\min}^{B}(r,t)/(3.0\,\text{\AA})\right],
& \text{if an interaction of type $t$ is present},\\
0, & \text{otherwise},
\end{cases}
\]
and compute
\[
\operatorname{IFA}(P,L)
=
\sum_{r:\,\|\mathbf u_P(r)\|+\|\mathbf u_L(r)\|>0}
\cos\!\left(\mathbf u_P(r),\mathbf u_L(r)\right).
\]
Here $d_{\min}^{B}(r,t)$ denotes the closest interaction of type $t$ between
binder $B$ and residue $r$; repeated interactions retain the minimum distance.
IFA is validated against target-residue overlap on held-out pairs, showing strong agreement and supporting its reliability as an interface-similarity metric (Appendix~\ref{sec:supp_ifa_validation}).

We report the \textbf{Vina docking score (kcal/mol)} for docking quality \citep{eberhardt2021vina}, \textbf{Validity (\%)} as the proportion of generated molecules identified as chemically valid by RDKit, \textbf{Uniqueness (\%)} as the proportion of unique generated molecules, and \textbf{Diversity} as the mean pairwise Tanimoto distance between ECFP4 fingerprints \citep{rogers2010ecfp}. We additionally report \textbf{AiZynth.} as the AiZynthFinder success rate, and \textbf{PoseCheck} as the summed protein--ligand interaction count in the docked pose. To ensure a fair aggregate comparison, PoseCheck results are reported only for models with successful evaluation on all 10 targets.

\subsection{Transferring Peptide-Target Interaction Patterns to Small Molecules}
\label{sec:encoder_experiments}
\begin{figure*}[!h]
\centering
\includegraphics[width=0.90\textwidth]{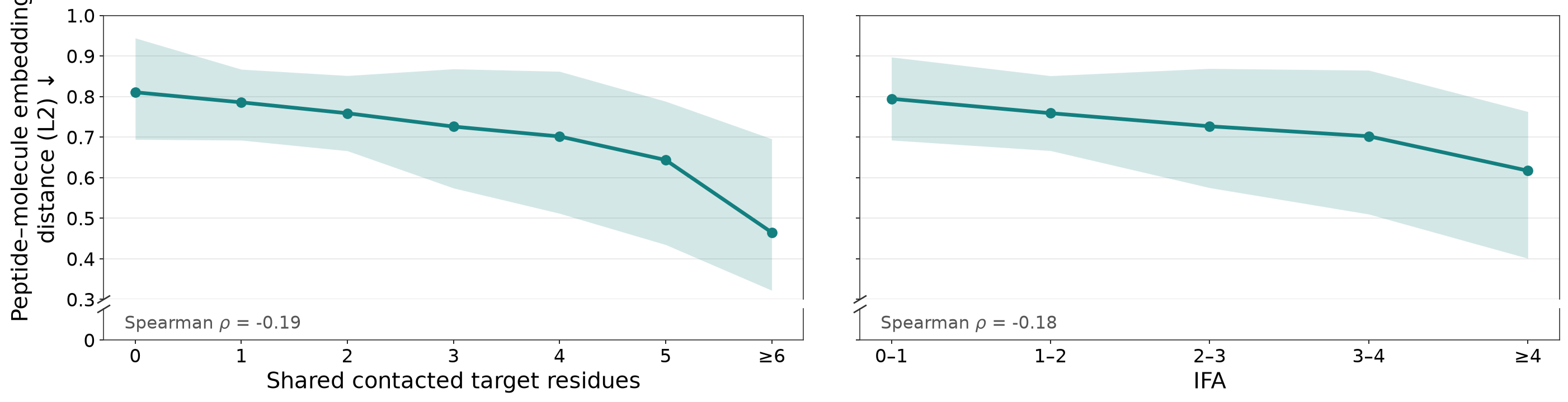}
\caption{\textbf{Peptide--molecule embedding distance across target-residue overlap and IFA.} \emph{Left:} Embedding distance decreases as shared contacted target residues increase, clearly distinguishing overlap levels. \emph{Right:} Higher IFA shows the same trend. Lines indicate the median and shaded regions the interquartile range (Q1--Q3).}
\label{fig:encoder}
\vspace{-0.6em}
\end{figure*}

We evaluate whether the learned cross-modal interface encoder captures interaction-pattern similarity by relating peptide--molecule embedding distance to target-residue overlap and IFA. Across held-out pairs, stronger interface agreement is associated with smaller embedding distances (Figure~\ref{fig:encoder}).

\subsection{Generating PPI Inhibitor with Peptide-Target Interface Guidance}
\label{sec:guidance_experiments}

\begin{figure*}[t]

\noindent\textbf{(a)}\par
\vspace{0.2em}
\includegraphics[width=\textwidth]
{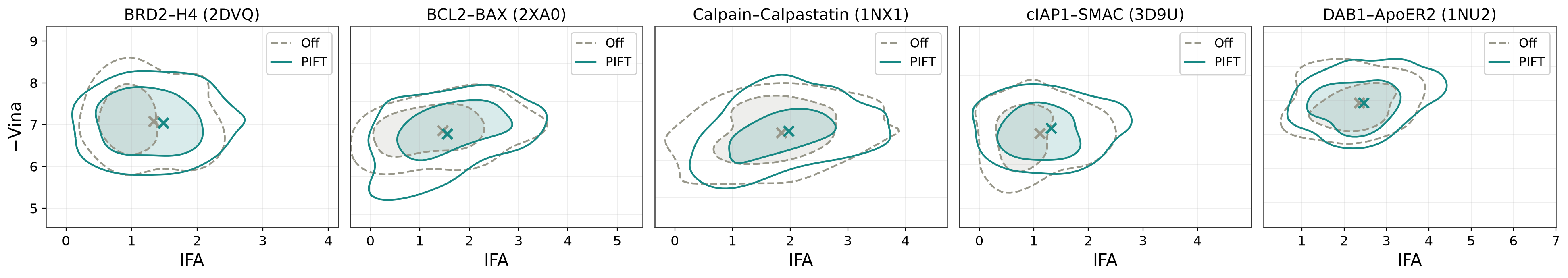}

\vspace{-0.9em}
\centering
\noindent
\begin{minipage}[t]{0.48\textwidth}%
\vspace{0pt}
\raggedright

\textbf{(b)}\par
\vspace{0.2em}

\refstepcounter{table}\label{tab:guidance_ablation}

\centering
\normalsize
\setlength{\tabcolsep}{1.2pt}
\renewcommand{\arraystretch}{1.18}

\resizebox{0.96\linewidth}{!}{%
\begin{tabular}{@{}lcccccc@{}}
\toprule
Guidance
& IFA $\uparrow$
& Precision $\uparrow$
& Res. count
& Vina $\downarrow$
& Success $\uparrow$ \\
\midrule
Off
& 1.975 & 0.448 & 3.360 & -7.662 & 25.3\% \\
Coordinate
& 2.016 & 0.444 & 3.439 & \textbf{-7.708} & 26.5\% \\
Discrete
& 2.016 & 0.457 & 3.425 & -7.691 & 24.7\% \\
\textbf{Joint}
& \textbf{2.049}
& \textbf{0.460}
& 3.385
& \underline{-7.703}
& \textbf{29.3\%} \\
\bottomrule
\end{tabular}%
}

\end{minipage}%
\hspace{0.005\textwidth}%
\begin{minipage}[t]{0.50\textwidth}%
\vspace{0pt}
\raggedright

\textbf{(c)}\par
\vspace{-0.15em}

\centering
\includegraphics[width=0.78\linewidth]
{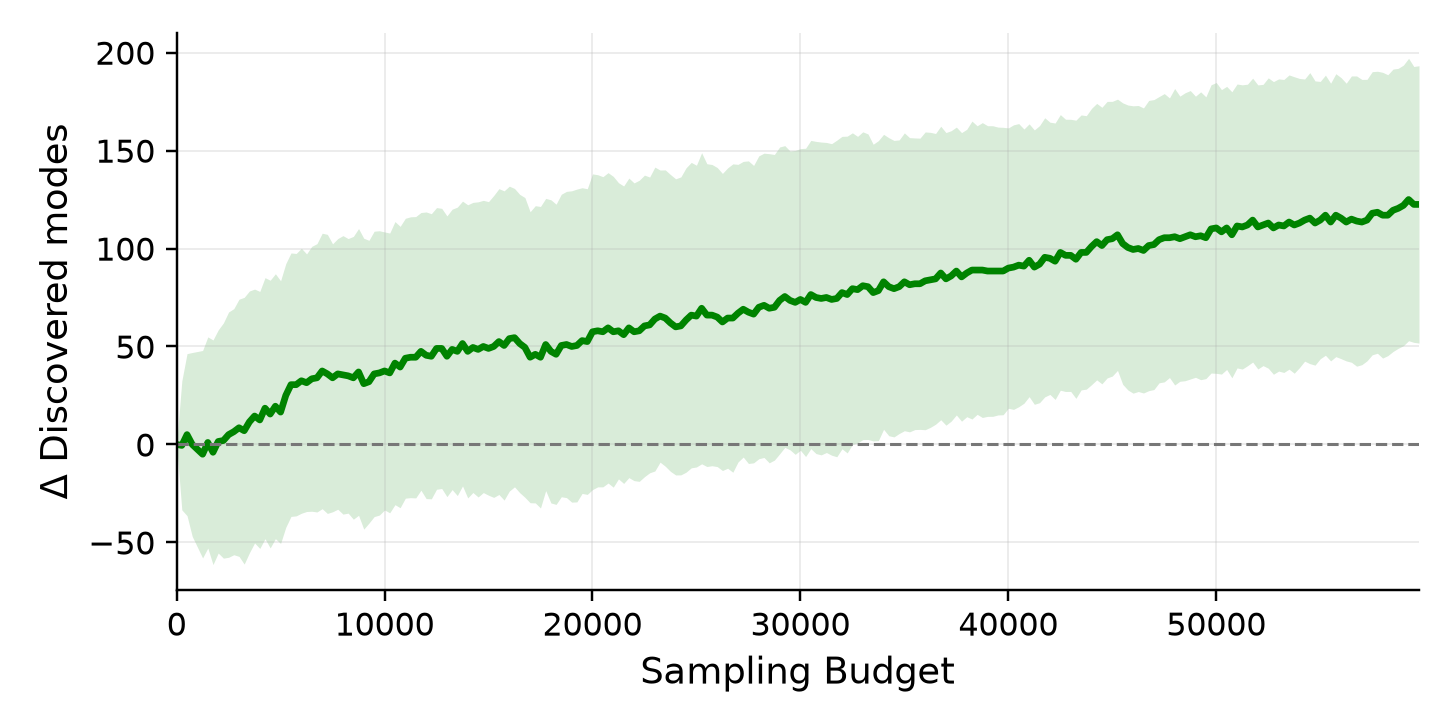}

\label{fig:guidance-yield}
\end{minipage}

\vspace{-0.9em}

\begin{subfigure}[b]{0.96\textwidth}
\textbf{(d)}\par
\vspace{0.2em}

\centering
\includegraphics[width=\linewidth]
{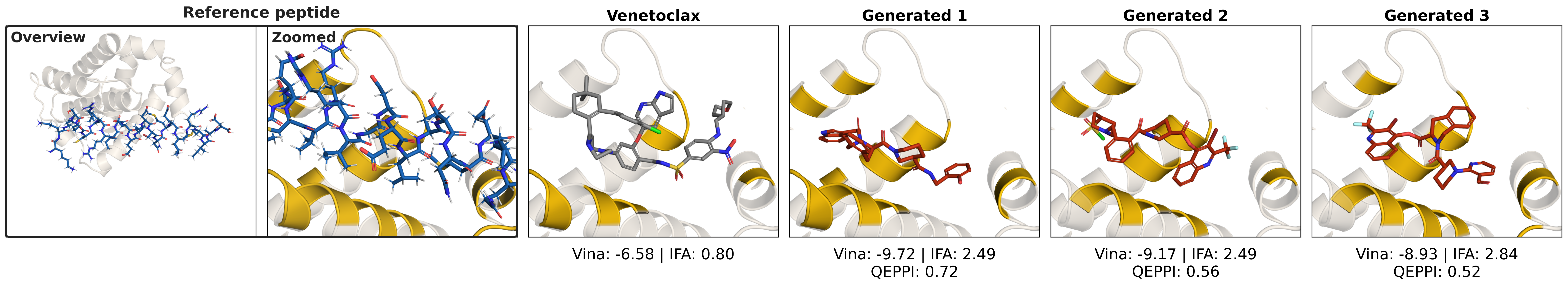}

\label{fig:guidance-structure}
\end{subfigure}

\caption{\textbf{Peptide-reference joint guidance.} \textbf{(a) PIFT and Off populations in IFA--$-$Vina space.} Shared density regions show that guidance largely preserves
the distribution supported by the DTM-adapted prior, while guided contours
extend toward higher interface agreement. Full results for the 10 PPI targets are provided in
Appendix~\ref{sec:supp_guidance_full}.\textbf{(b) Joint guidance ablation.}
Precision measures agreement of generated and reference
$(\text{residue},\text{interaction type})$ contacts; success requires both IFA
above and Vina below the target-specific Off median. \textbf{(c) Fixed-budget discovery.} The curve reports the cumulative
number of additional qualified modes discovered by PIFT relative to Off; the line is the median and
shading the approximate 95\% interval. \textbf{(d) 3D illustration of BCL2--BAX reference peptide, Venetoclax, and guided
molecules} Generated molecules demonstrate chemically distinct realizations of the reference
interface.}
\label{fig:guidance}
\vspace{-0.6em}
\end{figure*}

\paragraph{Overall population shift.}
We first assess whether peptide-reference guidance shifts generation toward molecules that better recover the desired interface while preserving the binding quality of the adapted prior. Relative to Off, guidance largely preserves the Vina distribution while extending sampling into higher-IFA regions, enabling discovery of molecules with stronger peptide-interface agreement (Figure~\ref{fig:guidance}a).

\paragraph{Joint guidance ablation.}
To evaluate the effect of joint guidance, we ablate the discrete and continuous geometric branches (Figure~\ref{fig:guidance}b). Coordinate guidance achieves the strongest single-branch docking score and the highest target-residue coverage, whereas discrete guidance yields higher interface precision despite contacting fewer residues. These differences suggest complementary roles: coordinate updates improve the spatial realization of the molecule at the interface, while discrete updates favor building blocks with more reference-compatible interaction functionality. Joint guidance combines both effects, achieving the highest IFA, precision, and success rate, while retaining near-best docking quality without unnecessarily expanding residue coverage.

\paragraph{Fixed-budget discovery.} 
We evaluate discovery under matched sampling budgets for each target, defining a qualified mode as IFA $>1.5$, QEPPI $>0.5$, Vina $<-6$ kcal/mol, and Tanimoto similarity $<0.5$ to any previously discovered mode.\footnote{For IL2, we use Vina $<-4$ kcal/mol to identify a comparable number of discovered modes relative to the other targets.}. Figure~\ref{fig:guidance}c reports the number of qualified modes across targets relative to Off. The increase shows that peptide-reference guidance more effectively discovers molecules that jointly satisfy interface compatibility, docking quality, and diversity criteria.

\paragraph{Reference specificity.}
We evaluate reference specificity on MDM2 using two experimentally resolved peptide binders, 1YCR and 3LNJ, whose target-contact patterns partially overlap but remain distinguishable. The results show clear reference-dependent guidance, indicating that the model responds to the selected peptide interaction pattern rather than simply favoring generic MDM2-binding features. Detailed results are provided in Appendix~\ref{sec:supp_reference_specificity}.

\subsection{Molecular Quality and Binding Performance}
\label{sec:quality}

\paragraph{External baseline comparison.}

Across individual PPI targets, PIFT consistently outperforms or remains competitive with diverse generative baselines, achieving the best mean and median Vina scores ($-7.68$/$-7.83$, Table~\ref{tab:vina}) and the highest aggregate IFA ($1.993$, Table~\ref{tab:molecular-properties}). Notably, PIFT shows a clear advantage over diffusion- and flow-based baselines that degrade sharply on difficult PPI targets. Both diffusion's stochastic denoising and flow matching's deterministic transport rely on a localized ligand-sized region defined by receptor geometry, making their conditioning less reliable when no such cavity is well defined. This vulnerability is apparent on BCL2--BAX, where TargetDiff and Peptide2Mol fail to generate usable molecules and FLOWR becomes structurally degenerate, while PIFT reaches the best Vina score ($-8.22$) among all baselines. We attribute this robustness to conditioning on the peptide-defined cross-modal interaction pattern rather than relying solely on receptor geometry. Beyond docking, our model achieves the highest aggregate IFA and PoseCheck interaction count, while retaining perfect uniqueness and a high fraction of chemically valid molecules.
\begin{table*}[!h]
\caption{\textbf{Vina docking performance.} Average Vina docking score for each targets, and the average and median values across 10 PPI targets. Results for remaining 5 targets are reported in Appendix \ref{sec:supp_guidance_full}. Best results are in bold.}
\label{tab:vina}
\centering
\scriptsize
\renewcommand{\arraystretch}{1.08}
\resizebox{\textwidth}{!}{%
\begin{tabular}{C{1.70cm}crrrrrrr}
\toprule
& & \multicolumn{5}{c}{Average Vina Docking Score (kcal/mol, $\downarrow$)} & \multicolumn{2}{c}{} \\
\cmidrule{3-7}\cmidrule{8-9}
Category & Model & \raisebox{-0.4ex}{\shortstack[c]{BRD2--H4}} & \raisebox{-0.4ex}{\shortstack[c]{BCL2--BAX}} & \raisebox{-0.4ex}{\shortstack[c]{Calpain--Calpastatin}} & \raisebox{-0.4ex}{\shortstack[c]{cIAP1--SMAC}} & \raisebox{-0.4ex}{\shortstack[c]{DAB1--ApoER2}} & \raisebox{-0.4ex}{Avg.} & \raisebox{-0.4ex}{Med.} \\
\midrule
\multirow{1}{*}{Autoregressive} & Hot2Mol & \vina{-7.00}{0.04} & \vina{-7.61}{0.05} & \vina{-6.56}{0.03} & \vina{-6.59}{0.05} & \vina{-7.35}{0.02} & -7.01 & -7.06 \\
\midrule
Search-based & SynGA & \vina{-6.47}{0.09} & \vina{-7.33}{0.05} & \vina{-6.50}{0.03} & \vina{-6.04}{0.09} & \vina{-6.78}{0.03} & -6.62 & -6.64 \\
\midrule
GFlowNet-based & CGFlow & \vinab{-7.20}{0.05} & \vina{-7.93}{0.03} & \vina{-6.91}{0.03} & \vina{-6.38}{0.03} & \vina{-6.99}{0.05} & -7.03 & -7.10 \\
\midrule
\multirow{6}{*}{\begin{tabular}[c]{@{}c@{}}Diffusion /\\ Flow-based\end{tabular}} & TargetDiff & \vina{-5.30}{0.04} & -- & \vina{-6.74}{0.03} & \vina{-3.50}{0.03} & \vina{-6.02}{0.08} & -6.27 & -6.82 \\
& MolCRAFT & \vina{-5.02}{0.02} & \vina{-6.84}{0.03} & \vina{-7.50}{0.04} & \vina{-3.90}{0.04} & \vina{-5.32}{0.01} & -5.81 & -6.25 \\
& Flowr & \vina{-4.00}{0.19} & \vina{-1.25}{0.00}\textsuperscript{\dag} & \vina{-5.00}{0.08} & \vina{-6.39}{0.05} & \vina{-6.04}{0.21} & -5.74 & -5.91 \\
& SynCoGen & \vina{-6.68}{0.08} & \vina{-7.70}{0.04} & \vina{-7.00}{0.05} & \vina{-6.33}{0.02} & \vina{-7.05}{0.08} & -6.92 & -7.00 \\
& Peptide2Mol & \vina{-5.68}{0.54} & -- & \vina{-6.18}{0.33} & \vina{-5.12}{0.23} & \vina{-4.90}{0.07} & -5.51 & -5.59 \\
\cmidrule{2-9}
& \textbf{PIFT} & \vina{-6.99}{0.04} & \vinab{-8.22}{0.06} & \vinab{-7.75}{0.06} & \vinab{-6.78}{0.05} & \vinab{-7.92}{0.04} & \textbf{-7.68} & \textbf{-7.83} \\
\bottomrule
\end{tabular}%
}
\vspace{0em}

\end{table*}

\begin{table*}[!h]
\vspace{-1.0em}
\caption{\textbf{Molecular-generation quality.} Best results are in bold.}
\label{tab:molecular-properties}
\centering
\tiny
\renewcommand{\arraystretch}{1.02}
\resizebox{0.88\textwidth}{!}{%
\begin{tabular}{C{1.65cm}ccccccc}
\toprule
Category & Model & IFA $\uparrow$ & Validity $\uparrow$ & AiZynth. $\uparrow$ & Uniqueness $\uparrow$ & Diversity $\uparrow$ & PoseCheck $\uparrow$ \\
\midrule
Autoregressive & Hot2Mol & 1.347 & 0.975 & 0.529 & \textbf{1.000} & 0.819 & 10.485 \\
\midrule
Search-based & SynGA & 1.329 & \textbf{1.000} & 0.258 & \textbf{1.000} & 0.878 & 11.071 \\
\midrule
GFlowNet-based & CGFlow & 1.182 & 0.978 & 0.575 & 0.972 & 0.828 & 9.193 \\
\midrule
\multirow{6}{*}{\begin{tabular}[c]{@{}c@{}}Diffusion /\\ Flow-based\end{tabular}} & TargetDiff & 1.227 & 0.571 & 0.106 & \textbf{1.000} & 0.891 & --- \\
& MolCRAFT & 1.420 & 0.864 & 0.226 & 0.976 & 0.854 & 11.340 \\
& Flowr & 1.290 & 0.909 & 0.023 & \textbf{1.000} & 0.842 & --- \\
& SynCoGen & 1.384 & 0.957 & \textbf{0.643} & \textbf{1.000} & 0.804 & 10.648 \\
& Peptide2Mol & 1.068 & 0.088 & 0.011 & \textbf{1.000} & \textbf{0.893} & --- \\
\cmidrule{2-8}
& \textbf{PIFT} & \textbf{1.993} & 0.947 & \underline{0.612} & \textbf{1.000} & 0.709 & \textbf{12.351} \\
\bottomrule
\end{tabular}%
}
\end{table*}

\section{Conclusion}

In this work, we introduced PIFT, a novel peptide-guided framework for small-molecule PPI inhibitor generation. 
PIFT effectively represents peptide-target and small-molecule-target interaction patterns in a shared interface space, and steers the molecular prior toward peptide-compatible regions.

Across diverse and challenging PPI targets, PIFT identifies peptide-compatible molecules with strong docking performance while retaining predicted synthesizability and favorable molecular properties.
These results demonstrate PIFT's ability to effectively discover diverse successful molecular candidates for PPI interfaces where conventional 3d molecular generation models struggle.

PIFT thus offers a new route toward the long-standing challenge of small-molecule PPI inhibitor design. Beyond peptide-to-small-molecule transfer, future work should extend this principle to interaction-guided generation across diverse biomolecular modalities.

\subsection*{AI use statement}

In this work, we used generative AI tools to polish the writing by refining grammar. 
We also used these tools to search for relevant literature.
No AI-generated text or suggestions were accepted without verification and editing by the authors. 
We did not use generative AI tools for research ideation or execution, drafting sections of the paper, generating synthetic datasets or proving mathematical claims.
We take responsibility for the final content of this work, including any text, claims, or artifacts produced with the aid of generative AI.

\bibliographystyle{iclr2027_conference}
\bibliography{iclr2027_conference}

\newpage
\appendix

\begin{center}
\Large\textbf{Appendix}
\end{center}
\vspace{0.5em}

\section{Training Details}
\label{app:training}

\subsection{Interface Encoder}

\begin{figure*}[!h]
\centering
\includegraphics[width=0.3\textwidth]{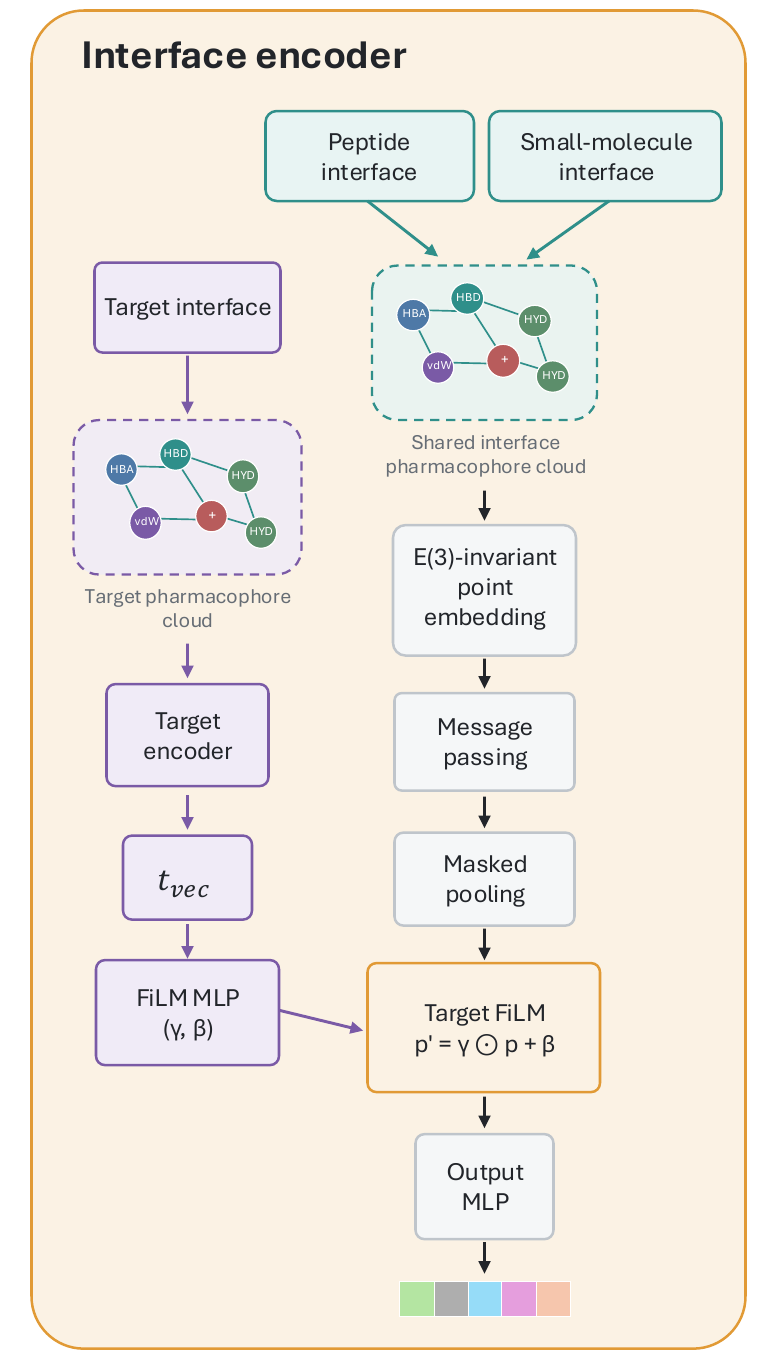}
\caption{\textbf{Target-conditioned Cross-Modal Interface encoder architecture.}}
\label{fig:appendix-encoder}
\end{figure*}

We provide a schematic of the target-conditioned cross-modal interface encoder in Figure~\ref{fig:appendix-encoder}.

\subsubsection{Dataset Construction}
\label{app:dataset}

We apply the following filtering, deduplication, partitioning, and query-construction procedures to the four structural resources described in Section~\ref{sec:dataset}.

Candidates are merged across sources and filtered by: (i) at least four target-contact residues; (ii) for small molecules, at least four heavy atoms and one ProLIF-detected interaction type; (iii) for the lower-confidence source, a Vina score at most $-6.0$, at least four eligible poses per interface, and retention of the top $25\%$ by score; (iv) an interface point cloud (Section~\ref{sec:pharmacophore}) containing at least four points in the shared frame; and (v) cross-source deduplication, retaining the higher-confidence instance when sources agree. We additionally cap cloud size at the train-split $99.5$th percentile: 237 points for peptides and 37 for small molecules.

We partition data at the UniProt-accession level. Accessions are grouped into connected components under pairwise GRIM pharmacophore similarity \citep{desaphy2013grim} at threshold $0.80$, then assigned to a fixed-seed $80/20$ train/test split; no accession occurs in both partitions. The resulting corpus contains 3,090 peptide instances from 271 accessions (2,578 train; 512 test) and 97,669 small-molecule instances from 292 accessions (73,494 train; 24,175 test). Restricting representation training to examples with extractable target clouds gives 2,467 peptide and 72,919 small-molecule instances across 215 shared accessions.

For target-conditioned alignment, a peptide query is eligible when it has a real target cloud and at least four same-target positive and four negative molecular endpoints. Labels are computed from contact profiles in the same curated pool: a positive endpoint has nonzero target-contact and interaction-type agreement with the query; a negative has zero agreement and a pocket-centroid distance below 15~\AA{} in the shared frame. This produces 82 eligible queries across 18 targets. Held-out evaluation uses the disjoint accession-level test partition, further restricted to pairs with extractable target clouds and usable interaction profiles.

\subsubsection{Hyperparameters}

Here we summarize the default hyperparameters used for training the interface encoder.

\begin{table}[!h]
\centering
\small
\caption{\textbf{Default hyperparameters.}}
\label{tab:encoder-hyperparameters}
\begin{tabular}{C{0.64\linewidth}c}
\toprule
Hyperparameters & Values \\
\midrule
Interface encoder hidden dimension & 128 \\
Interface embedding dimension & 64 \\
Target encoder & hidden dim. 64, output dim. 32 \\
\midrule
\multicolumn{2}{l}{\textbf{Stage I: modality-shared geometric pretraining}} \\
Epochs & 30 \\
Steps per epoch & 2,000 \\
Batch size & 128 \\
Learning rate & $1\times10^{-4}$ \\
Weight decay & $1\times10^{-4}$ \\
Partial-view margin $m_p$ & 1.0 \\
\midrule
\multicolumn{2}{l}{\textbf{Stage II: target-conditioned cross-modal alignment}} \\
Training cycles & 150 \\
Queries per rank & 2 \\
Learning rate & $3\times10^{-5}$ \\
Min. positive / hard examples & 4 / 4 \\
Examples per label & 4 \\
Max. queries per target & 6 \\
Ranking margin & 0.1 \\
Temperature & 0.1 \\
$\gamma$-regularization weight & 0.2 \\
Target-vector diversity weight & 1.0 \\
Target-vector diversity margin & 0.3 \\
\bottomrule
\end{tabular}
\end{table}

\subsection{PPI-Oriented Prior Adaptation}
\label{app:dtm}

For target-specific prior adaptation, let $V(x)$ be AutoDock Vina score, $Q(x)$ QEPPI score \citep{kosugi2021}, and $h(x)$ heavy-atom count. Positive reward used in implementation is
\begin{align}
R_{\mathrm{PPI}}(x)&=\max(-0.1V(x),10^{-5})Q(x)S_{\mathrm{size}}(h(x)),\\
S_{\mathrm{size}}(h)&=
\begin{cases}
1,&h\le45,\\
\exp[-0.15(h-45)],&h>45.
\end{cases}
\label{eq:ppi-reward-app}
\end{align}
DTM uses log reward $r(x)=\log\max(R_{\mathrm{PPI}}(x),10^{-12})$, centered by current replay-pool mean before tilting. This yields target distribution $p_R(x)\propto p_0(x)\exp(\beta r(x))$.

For masked position $i$, let $\pi_a(\cdot\mid x_t,i)$ be frozen reference posterior from previous tilt phase, $\pi_\theta$ trainable posterior, $x_1^i$ completed rollout token, $h$ local tilt increment, and $c$ distillation coefficient. Reward-induced local target is
\begin{equation}
T_c(v\mid x_t,i,x_1)=
\frac{c\pi_a(v\mid x_t,i)+[\exp(hr(x_1))-c]\mathbf 1[v=x_1^i]}
{\exp(hr(x_1))}.
\end{equation}
Corresponding weighted cross-entropy can be evaluated without normalizing sequence-level tilted likelihood:
\begin{equation}
\exp(hr)\,\mathrm{CE}(T_c,\pi_\theta)
=c\,\mathrm{CE}(\pi_a,\pi_\theta)
-[\exp(hr)-c]\log\pi_\theta(x_1^i\mid x_t,i).
\end{equation}
Same objective is applied to masked node and edge decisions alongside original coordinate-flow objective. This adaptation changes generic synthesis-aware prior into target-specific PPI prior but uses no peptide reference.

\subsection{Joint guidance with peptide reference}
\subsubsection{Hyperparameters}
Here we provide the hyperparameters used for the PIFT sampling procedure.

\begin{table*}[!h]
\centering
\small
\caption{\textbf{Default hyperparameters.}}
\label{tab:guidance-hyperparameters}
\resizebox{0.92\textwidth}{!}{%
\begin{tabular}{C{0.39\textwidth}C{0.25\textwidth}C{0.25\textwidth}}
\toprule
Hyperparameters & Discrete branch & Coordinate branch \\
\midrule
Guidance start $t_{\mathrm{start}}$ & 0.5 & 0.5 \\
Schedule exponent $p$ & 2.0 & 2.0 \\
Gradient clipping & 5.0 & 5.0 \\
Guidance scale $\gamma_{\max}$ & 0.1 & 0.1 \\
Margin fraction & 1.2 & --- \\
ST-Gumbel MC worlds $M$ & 4 & --- \\
ST-Gumbel temperature $\tau$ & 1.0 & --- \\
\bottomrule
\end{tabular}%
}
\end{table*}

\subsubsection{Gradient Stabilization}
\label{app:gradient-stabilization}

The magnitude and direction of the interface-loss gradient can vary substantially
across reverse-sampling steps as the molecular state becomes progressively
resolved. Applying these raw gradients directly can therefore produce abrupt
corrections that perturb the pretrained sampling trajectory. We stabilize the
guidance signal using RMS normalization and clipping to control its instantaneous
scale, followed by Adam-style moment smoothing to reduce step-to-step variation.

For a preprocessed gradient $\tilde{\mathbf g}_t$, we maintain
\begin{equation}
\begin{aligned}
\mathbf m_t
&=\beta_1\mathbf m_{t-1}+(1-\beta_1)\tilde{\mathbf g}_t,\\
\mathbf v_t
&=\beta_2\mathbf v_{t-1}+(1-\beta_2)\tilde{\mathbf g}_t^{\odot 2},\\
\hat{\mathbf g}_t
&=\frac{\hat{\mathbf m}_t}
{\sqrt{\hat{\mathbf v}_t}+\epsilon},
\end{aligned}
\label{eq:gradient_stabilization}
\end{equation}
where $\hat{\mathbf m}_t$ and $\hat{\mathbf v}_t$ are bias-corrected moments.
The stabilized direction $\hat{\mathbf g}_t$ is then used in the corresponding
logit or coordinate update.

\newpage
\section{Experimental Details}
\label{app:experiments}

\subsection{Baselines}
\label{app:baselines}
\label{sec:baseline_details}

Below, we summarize the core generative mechanisms and conditioning strategies of the baseline methods used in our evaluation.

For pocket-conditioned 3D generation, \textbf{TargetDiff} \citep{guan2023targetdiff} uses $\mathrm{SE}(3)$-equivariant diffusion to jointly generate ligand atom types and 3D coordinates conditioned on protein-pocket atoms, whereas \textbf{MolCRAFT} \citep{qu2024molcraft} models molecular geometry and discrete chemical attributes in a unified continuous parameter space using a pocket-conditioned Bayesian Flow Network. \textbf{FLOWR} \citep{cremer2026flowr} combines continuous coordinate flow matching with categorical atom and bond flows and integrates a separately encoded protein pocket through equivariant cross-attention; it can additionally condition on desired protein--ligand interaction features.

For synthesis-aware generation, pocket-specific \textbf{CGFlow} \citep{shen2025cgflow} jointly searches over reaction-based molecular construction and 3D binding poses within the target complex using compositional generative flows that combine a GFlowNet for molecular construction with flow matching for coordinates, while \textbf{SynCoGen} \citep{rekesh2026syncogen} couples absorbing-state discrete diffusion over graph-structured building-block and reaction representations with continuous flow matching over coordinates. \textbf{SynGA} \citep{lo2026synga} instead performs target-specific genetic optimization directly over synthesis routes using synthesizability-preserving crossover and mutation operators.

For PPI-specific conditioning, \textbf{Hot2Mol} \citep{sun2026hot2mol} integrates a conditional transformer with a variational autoencoder to generate molecules, whereas \textbf{Peptide2Mol} \citep{he2025peptide2mol} conditions an $\mathrm{E}(3)$-equivariant diffusion model jointly on a peptide binder and its surrounding protein environment to generate peptide-informed 3D small molecules.

\subsection{Hyperparameters}
Here we summarize the default hyperparameters used for sampling experiments.

\begin{table}[!h]
\centering
\small
\caption{\textbf{Default hyperparameters.}}
\label{tab:sampling-hyperparameters}
\begin{tabular}{C{0.68\linewidth}c}
\toprule
Hyperparameters & Values \\
\midrule
Reverse sampling steps & 100 \\
Batch size & 64 \\
Samples per target & 256 \\
Docking & UniDock \\
Docking seeds & 5 \\
\bottomrule
\end{tabular}
\end{table}

\newpage
\section{Extended Results}
\subsection{IFA Validation}
\label{sec:supp_ifa_validation}

We validate Interface Agreement (IFA) against the overlap of active target
residues on held-out peptide--molecule pairs. Residue-set Jaccard similarity
correlates with mean IFA with pooled Pearson $r=0.987$ and query-wise Spearman
$\rho=0.963$. The compatibility-restricted IFA variant shows similarly strong
agreement ($r=0.996$, $\rho=0.985$), supporting IFA as a graded measure of
peptide--molecule interaction recovery.

\subsection{Interface Encoder Architecture Ablation}
\label{sec:supp_encoder_ablation}

We examine how target conditioning FiLM layer affects cross-modal organization in the learned interface space. We compare the target-agnostic encoder with concatenation-based conditioning, Early FiLM, and the Late-FiLM design used in our final model. For each held-out peptide query, we evaluate whether a compatible small-molecule interface is embedded closer than a mismatched interface and measure the corresponding positive--negative distance margin. As shown in Table~\ref{tab:encoder_ablation}, Late FiLM yields the highest pair-ranking accuracy and the largest pair margin among the tested variants.

\begin{table}[!h]
\centering
\small
\caption{\textbf{Ablation of target conditioning.} Pair ranking measures the fraction of peptide queries for which the compatible small-molecule interface is embedded closer than its mismatched counterpart. Pair margin is the difference between mismatched- and compatible-pair distances, $d_{\mathrm{neg}}-d_{\mathrm{pos}}$, where larger values indicate stronger separation.}
\label{tab:encoder_ablation}
\begin{tabular}{C{0.52\linewidth}cc}
\toprule
Encoder variant & Pair ranking $\uparrow$ & Pair margin $\uparrow$ \\
\midrule
No target conditioning & \vina{0.572}{0.029} & \vina{0.067}{0.032} \\
Concatenation & \vina{0.586}{0.030} & \vina{0.054}{0.056} \\
Early FiLM & \vina{0.579}{0.018} & \vina{0.044}{0.024} \\
\textbf{PIFT (Late FiLM)} & \vinab{0.598}{0.042} & \vinab{0.099}{0.064} \\
\bottomrule
\end{tabular}
\end{table}

\newpage
\subsection{Reference Specificity Analysis}
\label{sec:supp_reference_specificity}
\begin{figure*}[!h]
\centering
\begin{subfigure}[t]{0.30\textwidth}
\vspace{0pt}
\centering
\makebox[\linewidth][l]{\textbf{(a)}}\\[0.25em]
\includegraphics[width=\linewidth]{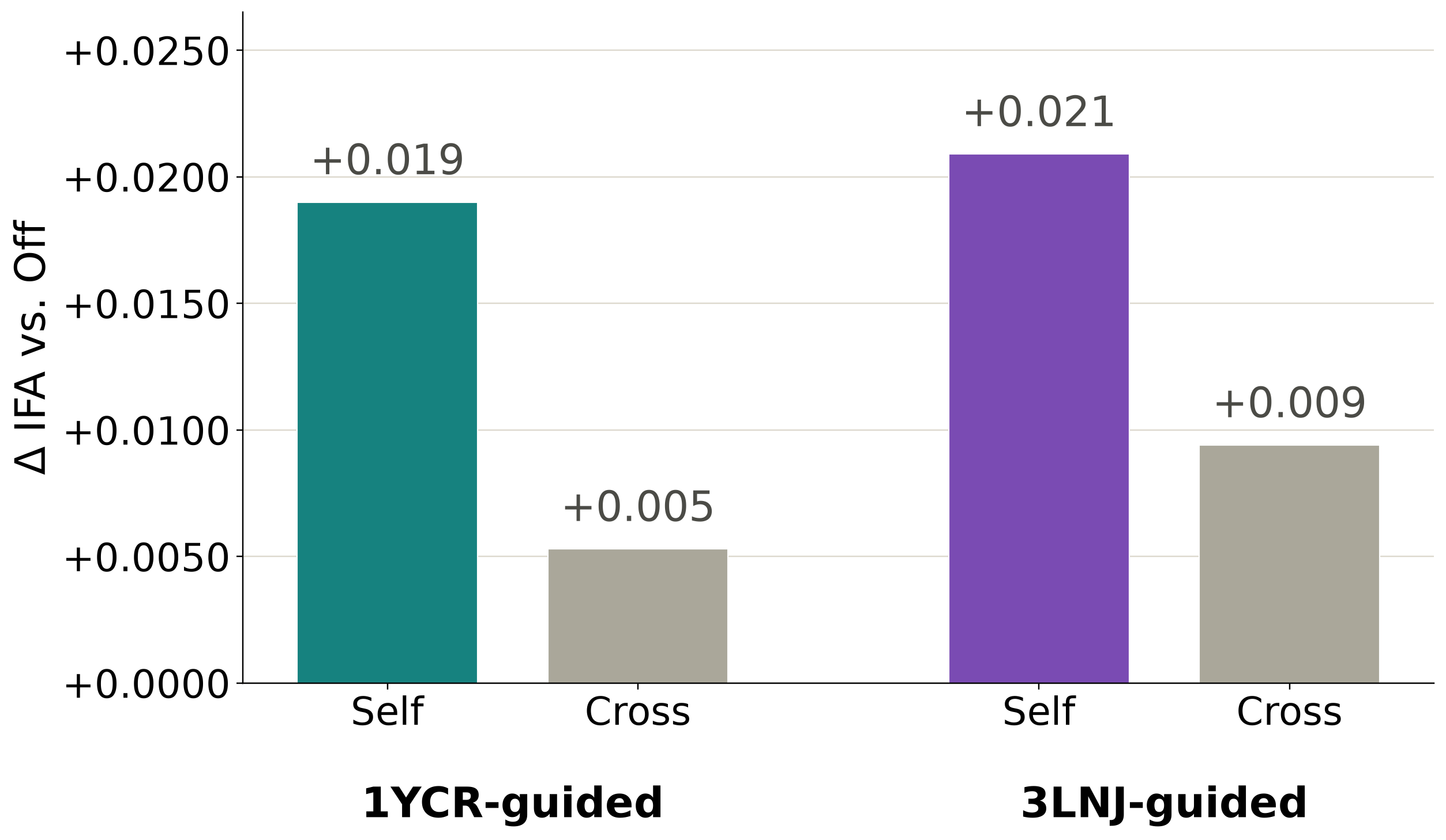}
\label{fig:reference-ifa}
\end{subfigure}\hspace{0.01\textwidth}
\begin{subfigure}[t]{0.67\textwidth}
\vspace{0pt}
\centering
\makebox[\linewidth][l]{\textbf{(b)}}\\[0.25em]
\includegraphics[width=\linewidth]{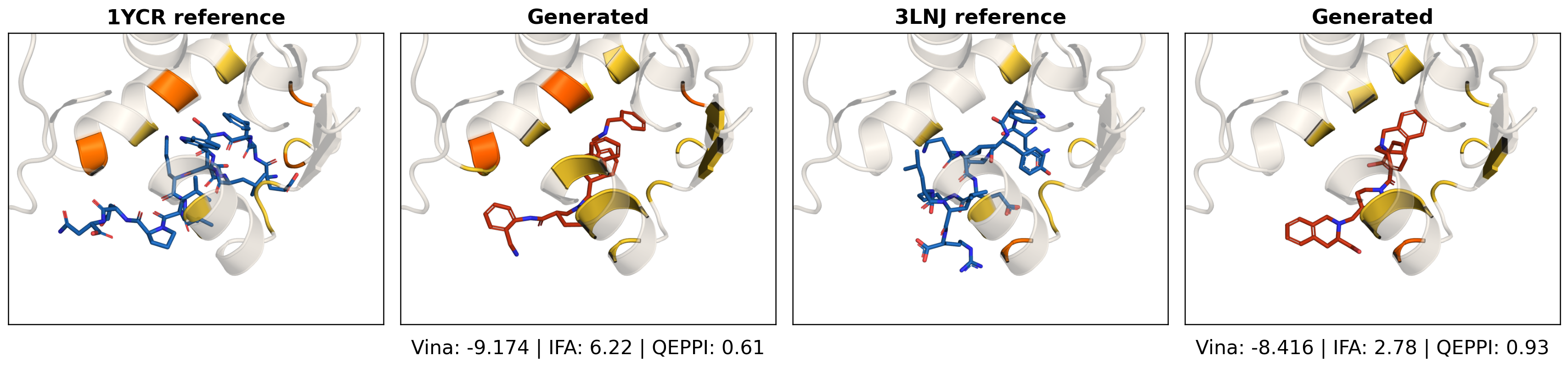}
\label{fig:reference-structure}
\end{subfigure}
\vspace{-0.05em}
\begin{subfigure}[t]{0.99\textwidth}
\vspace{0pt}
\centering
\makebox[\linewidth][l]{\textbf{(c)}}\\[0.25em]
\includegraphics[width=\linewidth]{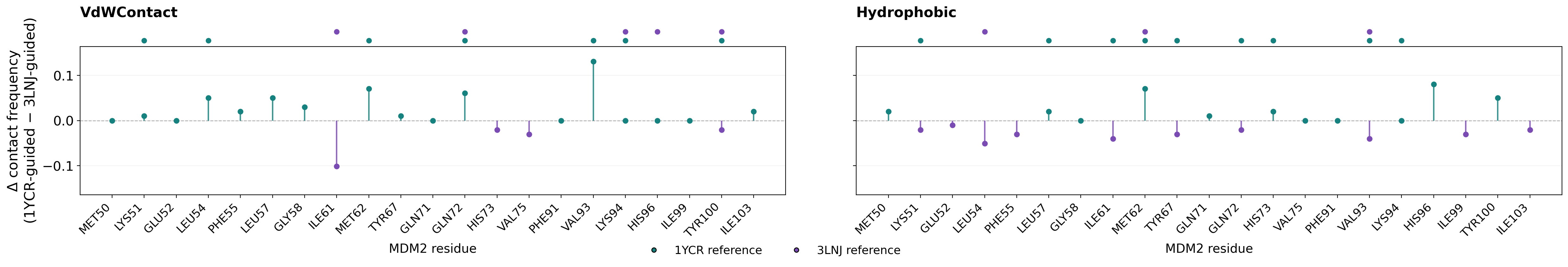}
\label{fig:reference-fingerprint}
\end{subfigure}
\vspace{-0.35em}
\captionsetup{skip=2pt}
\caption{\textbf{Reference-specificity of peptide guidance on MDM2.} (a) Self- and cross-reference IFA scoring. Changes in IFA relative to Off for the 1YCR- and 3LNJ-guided ensembles, evaluated against their own (self) or alternate (cross) peptide reference. Self-reference gains are larger for both guided ensembles. (b) Reference peptides and representative guided binding modes. Guided molecules occupy similar MDM2 regions while adopting reference-dependent local interaction geometries. (c) Differential population-level interaction fingerprints. Localized shifts reveal reference-dependent interaction patterns.}
\label{fig:reference_specificity}
\end{figure*}

We study reference-specific guidance behavior on MDM2 using two experimentally resolved peptide binders, 1YCR and 3LNJ, whose target-contact patterns partially overlap but remain distinguishable. Figure~\ref{fig:reference_specificity} summarizes reference-specific effects at the score, structural, and population levels. Each guided population is evaluated against both its own reference and the alternate peptide. Relative to the guidance-off population, 1YCR-guided sampling shows a larger IFA improvement when evaluated against 1YCR than against 3LNJ, while 3LNJ-guided sampling exhibits the converse pattern. This differential response indicates that guidance is sensitive to the selected peptide interaction pattern rather than merely promoting a generic MDM2-compatible interface. Representative binding modes provide a structural view of this effect: molecules generated from the two references occupy similar regions of MDM2 but adopt distinct local interaction geometries around the peptide-contacting surface. At the population level, differential residue-by-interaction fingerprints over the observed hydrophobic and van der Waals interaction channels reveal localized shifts between the two guided ensembles, further supporting reference-dependent modulation of the target interface.

\newpage
\subsection{Full Results for PPI targets}
\label{sec:supp_guidance_full}

\begin{figure*}[!h]
\centering
\includegraphics[width=\textwidth]{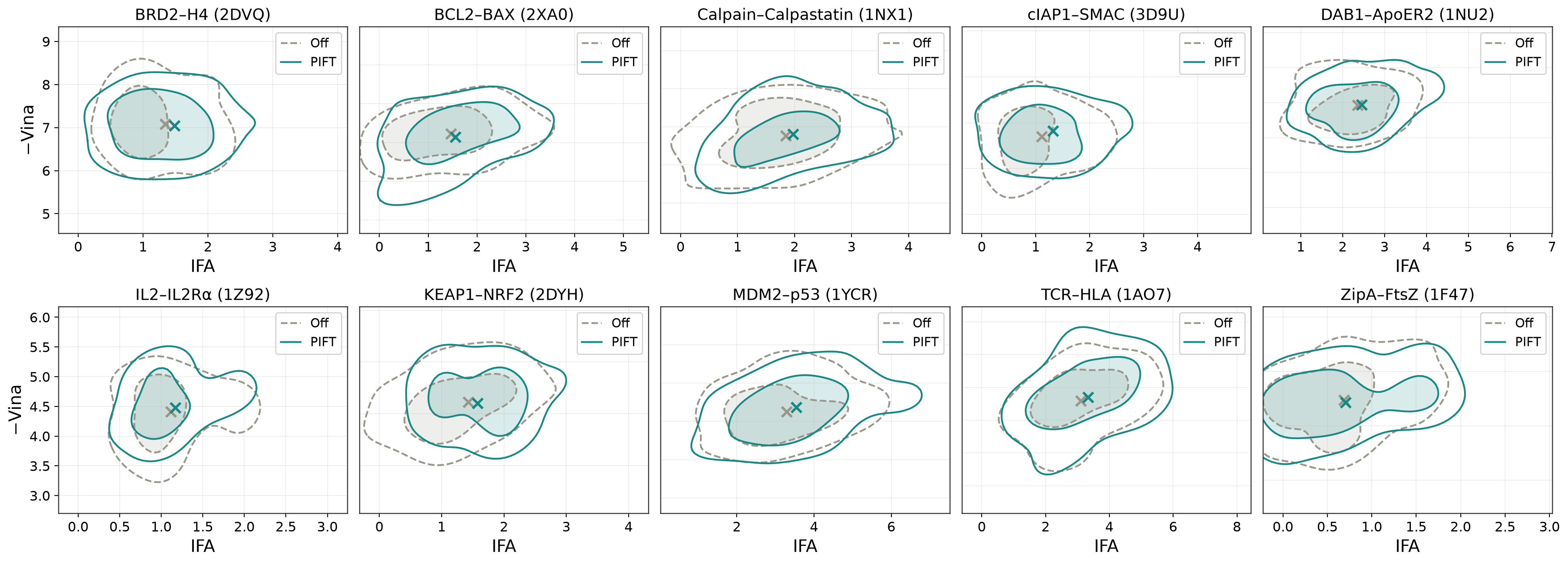}
\caption{\textbf{Full target-level results for guided and unguided molecular populations in IFA--$-$Vina space.}}
\label{fig:supp-ifa}
\end{figure*}

\begin{table*}[!h]
\centering
\scriptsize
\caption{\textbf{Average Vina docking score for 10 PPI targets.} Best results are in bold.}
\begin{tabular*}{\textwidth}{@{\extracolsep{\fill}}C{1.70cm}cccccc@{}}
\toprule
& & \multicolumn{5}{c}{Average Vina Docking Score (kcal/mol, $\downarrow$)} \\
\cmidrule{3-7}
Category & Model & \shortstack[c]{BRD2--H4} & \shortstack[c]{BCL2--BAX} & \shortstack[c]{Calpain--Calpastatin} & \shortstack[c]{cIAP1--SMAC} & \shortstack[c]{DAB1--ApoER2} \\
\midrule
Autoregressive & Hot2Mol & \vina{-7.00}{0.04} & \vina{-7.61}{0.05} & \vina{-6.56}{0.03} & \vina{-6.59}{0.05} & \vina{-7.35}{0.02} \\
\midrule
Search-based & SynGA & \vina{-6.47}{0.09} & \vina{-7.33}{0.05} & \vina{-6.50}{0.03} & \vina{-6.04}{0.09} & \vina{-6.78}{0.03} \\
\midrule
GFlowNet-based & CGFlow & \vinab{-7.20}{0.05} & \vina{-7.93}{0.03} & \vina{-6.91}{0.03} & \vina{-6.38}{0.03} & \vina{-6.99}{0.05} \\
\midrule
\multirow{6}{*}{\begin{tabular}[c]{@{}c@{}}Diffusion /\\ Flow-based\end{tabular}} & TargetDiff & \vina{-5.30}{0.04} & -- & \vina{-6.74}{0.03} & \vina{-3.50}{0.03} & \vina{-6.02}{0.08} \\
& MolCRAFT & \vina{-5.02}{0.02} & \vina{-6.84}{0.03} & \vina{-7.50}{0.04} & \vina{-3.90}{0.04} & \vina{-5.32}{0.01} \\
& Flowr & \vina{-4.00}{0.19} & \vina{-1.25}{0.00}\textsuperscript{\dag} & \vina{-5.00}{0.08} & \vina{-6.39}{0.05} & \vina{-6.04}{0.21} \\
& SynCoGen & \vina{-6.68}{0.08} & \vina{-7.70}{0.04} & \vina{-7.00}{0.05} & \vina{-6.33}{0.02} & \vina{-7.05}{0.08} \\
& Peptide2Mol & \vina{-5.68}{0.54} & -- & \vina{-6.18}{0.33} & \vina{-5.12}{0.23} & \vina{-4.90}{0.07} \\
\cmidrule{2-7}
& \textbf{PIFT} & \vina{-6.99}{0.04} & \vinab{-8.22}{0.06} & \vinab{-7.75}{0.06} & \vinab{-6.78}{0.05} & \vinab{-7.92}{0.04} \\
\bottomrule
\end{tabular*}

\vspace{0.8em}
\begin{tabular*}{\textwidth}{@{\extracolsep{\fill}}C{1.70cm}cccccc@{}}
\toprule
& & \multicolumn{5}{c}{Average Vina Docking Score (kcal/mol, $\downarrow$)} \\
\cmidrule{3-7}
Category & Model & \shortstack[c]{IL2--IL2R} & \shortstack[c]{KEAP1--NRF2} & \shortstack[c]{MDM2--p53} & \shortstack[c]{TCR--HLA} & \shortstack[c]{ZipA--FtsZ} \\
\midrule
Autoregressive & Hot2Mol & \vina{-4.35}{0.04} & \vina{-7.12}{0.05} & \vina{-7.54}{0.03} & \vina{-9.30}{0.01} & \vina{-6.69}{0.03} \\
\midrule
Search-based & SynGA & \vina{-4.04}{0.05} & \vina{-6.93}{0.04} & \vina{-6.94}{0.07} & \vina{-8.77}{0.06} & \vina{-6.37}{0.04} \\
\midrule
GFlowNet-based & CGFlow & \vina{-4.26}{0.04} & \vina{-7.48}{0.05} & \vina{-7.53}{0.04} & \vina{-8.86}{0.03} & \vina{-6.75}{0.04} \\
\midrule
\multirow{6}{*}{\begin{tabular}[c]{@{}c@{}}Diffusion /\\ Flow-based\end{tabular}} & TargetDiff & -- & \vina{-6.90}{0.04} & \vina{-6.95}{0.06} & \vina{-7.83}{0.06} & \vina{-6.94}{0.07} \\
& MolCRAFT & \vina{-3.30}{0.04} & \vina{-6.80}{0.05} & \vina{-6.96}{0.06} & \vina{-6.50}{0.10} & \vina{-5.99}{0.03} \\
& Flowr & \vinab{-4.61}{0.03} & \vina{-7.13}{0.02} & \vina{-7.93}{0.05} & \vina{-9.24}{0.03} & \vina{-5.77}{0.03} \\
& SynCoGen & \vina{-3.98}{0.05} & \vina{-7.46}{0.05} & \vina{-6.99}{0.04} & \vina{-9.33}{0.04} & \vina{-6.63}{0.06} \\
& Peptide2Mol & -- & \vina{-5.59}{0.07} & \vina{-5.60}{0.08} & -- & \vina{-5.47}{0.04} \\
\cmidrule{2-7}
& \textbf{PIFT} & \vina{-4.42}{0.05} & \vinab{-8.26}{0.06} & \vinab{-8.40}{0.06} & \vinab{-10.65}{0.06} & \vinab{-7.37}{0.05} \\
\bottomrule
\end{tabular*}
\end{table*}

\begin{figure*}[!h]
\centering
\includegraphics[width=0.96\textwidth]{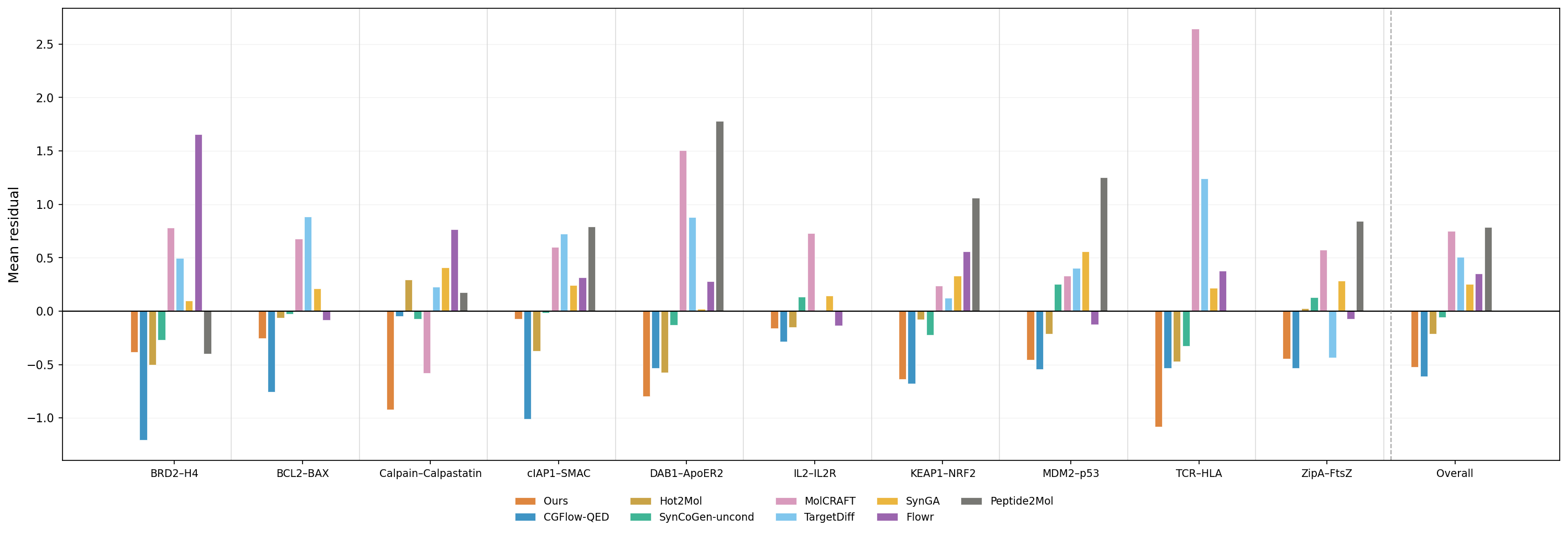}
\caption{\textbf{Heavy-atom-count-adjusted docking comparison.} Bars show mean residual Vina score after fitting a separate Vina--heavy-atom-count regression within each target; the final group averages residuals across all 10 targets. Lower residuals indicate more favorable docking scores than expected at matched molecular size.}
\label{fig:vina-hac}
\end{figure*}

\newpage
\paragraph{Docking efficiency.} To ensure that docking improvements are not solely driven by larger molecular size, we regress Vina score against heavy-atom count within each target and compare mean residuals across methods (Figure~\ref{fig:vina-hac}). PIFT remains favorable after size correction across most targets and achieves a strongly negative mean residual overall, indicating that its docking advantage does not rely on increasing molecular-size. Combined with the increased IFA, this supports selective transfer of peptide-defined interaction patterns rather than size-based mimicry.

\begin{table*}[!h]
\centering
\small
\caption{\textbf{PoseCheck metrics.} Average interaction counts for the 10-target benchmark, restricted to models with successful evaluation on all 10 targets. Values are means $\pm$ standard deviations over three runs; the main-text table reports only the corresponding Sum column. Best results are in bold, and the second-best H-bond acceptor value is underlined.}
\label{tab:posecheck}
\resizebox{\textwidth}{!}{%
\begin{tabular}{cccccc}
\toprule
& \multicolumn{5}{c}{PoseCheck Metrics ($\uparrow$)} \\
\cmidrule(lr){2-6}
Model & H-Bond Acceptors & H-Bond Donors & Van der Waals & Hydrophobic & Sum \\
\midrule
Hot2Mol & \vina{0.201}{0.004} & \vina{0.128}{0.008} & \vina{7.187}{0.044} & \vina{2.968}{0.046} & \vina{10.485}{0.066} \\
SynGA & \vina{0.241}{0.024} & \vina{0.071}{0.006} & \vina{7.797}{0.051} & \vina{2.961}{0.045} & \vina{11.071}{0.088} \\
CGFlow & \vina{0.130}{0.006} & \vina{0.054}{0.009} & \vina{6.583}{0.032} & \vina{2.427}{0.013} & \vina{9.193}{0.048} \\
MolCRAFT & \vina{0.232}{0.022} & \vinab{0.141}{0.024} & \vina{7.736}{0.093} & \vina{3.231}{0.044} & \vina{11.340}{0.124} \\
SynCoGen & \vinab{0.262}{0.012} & \vina{0.139}{0.008} & \vina{7.364}{0.041} & \vina{2.883}{0.037} & \vina{10.648}{0.071} \\
\textbf{PIFT} & \underline{0.251}\ {\tiny ($\pm$ 0.018)} & \vina{0.109}{0.007} & \vinab{8.115}{0.037} & \vinab{3.875}{0.040} & \vinab{12.351}{0.066} \\
\bottomrule
\end{tabular}%
}
\end{table*}

\end{document}